\documentclass[a4paper]{article}

\usepackage{INTERSPEECH_v2}

\usepackage{color}

\title{Multitask Learning with CTC and Segmental CRF for Speech Recognition}
\makeatother \name{ Liang Lu$^\dag$, Lingpeng Kong$^\ddag$, Chris Dyer$^\S$, and Noah A. Smith$^\diamond$}
\address{$^\dag$Toyota Technological Institute at Chicago, USA\\
  $^\ddag$School of Computer Science, Carnegie Mellon University, USA\\
 $^\S$Google DeepMind, London, UK \\
 $^\diamond$Computer Science \& Engineering,  University of Washington, Seattle, USA} 
\email{llu@ttic.edu, lingpenk@cs.cmu.edu, cdyer@google.com, nasmith@cs.washington.edu}

\begin{document}

\maketitle
\begin{abstract}
Segmental conditional random fields (SCRFs) and connectionist temporal classification (CTC) are two  sequence labeling methods used for end-to-end training of speech recognition models. Both models define a transcription probability by marginalizing decisions about latent segmentation alternatives to derive a sequence probability: the former uses a globally normalized joint model of segment labels and durations, and the latter classifies each frame as either an output symbol or a ``continuation'' of the previous label.  In this paper, we train a recognition model by optimizing an interpolation between the SCRF and CTC losses, where the same recurrent neural network (RNN) encoder is used for feature extraction for both outputs. We find that this multitask objective improves recognition accuracy when decoding with either the SCRF or CTC models. Additionally, we show that CTC can also be used to pretrain the RNN encoder, which improves the convergence rate when learning the joint model. 
%Segmental Conditional Random Field (SCRF) and Connectionist Temporal Classification (CTC) have been well studied for acoustic modeling. The former defines the loss function at the sequence level, while the latter defines the loss at the frame level. When combined with neural networks for feature extraction, both models can be trained end-to-end. In this paper, we study the joint training approach of the two models using an interpolated loss functions for speech recognition, where both SCRF and CTC share the same recurrent neural network (RNN) encoder used for feature extraction. We show that the joint training approach improves the recognition accuracy of both SCRF and CTC models due to the regularization effect, which is similar to the observation of HMMs where sequence-discriminative training using a sequence-level loss can benefit from the cross entropy (CE) based regularization. Moreover, we show that CTC can also be used to pretrain the RNN encoder, which improves the convergence speed of the joint model. 
\end{abstract}

\section{Introduction}

State-of-the-art speech recognition accuracy has significantly improved over the past few years since the application of deep neural networks~\cite{hinton2012deep, seide2011conversational}. Recently, it has been shown that with the application of both neural network acoustic model and language model, an automatic speech recognizer can approach  human-level accuracy on the Switchboard conversational speech recognition benchmark using around 2,000 hours of transcribed data~\cite{xiong2016}. While progress is mainly driven by well engineered neural network architectures and a large amount of training data, the hidden Markov model (HMM) that has been the backbone for speech recognition for decades is still playing a central role. Though tremendously successful for the problem of speech recognition, the HMM-based pipeline factorizes the whole system into several components, and building these components separately may be less computationally efficient when developing a large-scale system from thousands to hundred of thousands of examples~\cite{soltau2016neural}.

Recently, along with hybrid HMM/NN frameworks for speech recognition, there has been  increasing interest in end-to-end training approaches. The key idea is to directly map the input acoustic frames to output characters or words without the intermediate alignment to context-dependent phones used by HMMs. In particular, three architectures have been proposed for the goal of end-to-end learning:  connectionist temporal classification (CTC)~\cite{graves2014towards, Hannun2014Deep, sak2015learning, miao2015eesen}, sequence-to-sequence with attention model~\cite{chorowski2015attention, lu2015study, chan2016listen}, and neural network segmental conditional random field (SCRF)~\cite{abdel2013deep, lu2016segmental}. These end-to-end models simplify the pipeline of speech recognition significantly. They do not require intermediate alignment or segmentation like HMMs, instead, the alignment or segmentation is marginalized out during training for CTC and SCRF or inferred by the attention mechanism. In terms of the recognition accuracy, however, the end-to-end models usually lag behind their HMM-based counterparts. Though CTC has been shown to outperform HMM systems \cite{sak2015fast}, the improvement is based on the use of context-dependent phone targets and a very large amount of training data. Therefore, it has almost the same system complexity as HMM acoustic models. When the training data is less abundant, it has been shown that the accuracy of CTC systems degrades significantly~\cite{pundak2016lower}. 

However, end-to-end models have the flexibility to be combined to mitigate their individual weaknesses. For instance, multitask learning with attention models has been investigated for machine translation~\cite{luong2015multi}, and Mandarin speech recognition using joint Character-Pinyin training~\cite{chan2016online}. In~\cite{kim2016joint}, Kim et al. proposed a multitask learning approach to train a joint attention model and a CTC model using a shared encoder. They showed that the CTC auxiliary task can help the attention model to overcome the misalignment problem in the initial few epochs, and speed up the convergence of the attention model. Another nice property of the multitask learning approach is that the joint model can still be trained end-to-end. Inspired by this work, we study end-to-end training of a joint CTC and SCRF model using an interpolated loss function. The key difference of our study from~\cite{kim2016joint} is that the two loss functions of the CTC and attention models are locally normalized for each output token, and they are both trained using the cross entropy criterion. However, the SCRF loss function is normalized at the sequence-level, which is similar to the sequence discriminative training objective function for HMMs. From this perspective, the interpolation of CTC and SCRF loss functions is analogous to the sequence discriminative training of HMMs with CE regularization to overcome overfitting, where a sequence-level loss is also interpolated with a frame-level loss, e.g.,~\cite{su2013error}. Similar to the observations in~\cite{kim2016joint}, we demonstrate that the joint training approach improves the recognition accuracies of both CTC and SCRF acoustic models. Further, we also show that CTC can be used to pretrain the neural network feature extractor to speed up the convergence of the joint model. Experiments were performed on the TIMIT database. 

\section{Segmental Conditional Random Fields}
\label{sec:scrf}

SCRF is a variant of the linear-chain CRF model where each output token corresponds to a segment of input tokens instead of a single input instance. In the context of speech recognition, given a sequence of input vectors of $T$ frames ${\bm X} = ( {\bm x}_1, \cdots, {\bm x}_T )$ and its corresponding sequence of output labels ${\bm y} = ( y_1, \cdots, y_J)$, the zero-order linear-chain CRF defines the sequence-level conditional probability as
\begin{align}
\label{eq:crf}
P({\bm y} \mid {\bm X}) = \frac{1}{Z({\bm X})} \prod_{t=1}^T \exp f \left( y_t, {\bm x}_t \right),
\end{align}
where $Z({\bm X})$ denotes the normalization term, and $T=J$. Extension to higher order models is straightforward, but it is usually computationally much more expensive. The model defined in Eq.~\eqref{eq:crf} requires the length of ${\bm X}$ and ${\bm y}$ to be equal, which makes it inappropriate for speech recognition because the lengths of the input and output sequences are not equal. For the case where  $T\ge J$ as in speech recognition, SCRF defines the sequence-level conditional probability with the auxiliary segment labels ${\bm E} = ({\bm e}_1, \cdots, {\bm e}_J) $ as
\begin{align}
\label{eq:scrf}
P({\bm y}, {\bm E} \mid {\bm X}) = \frac{1}{Z({\bm X})} \prod_{j=1}^J \exp f \left( y_j, {\bm e}_j, \bar{\bm x}_j \right),
\end{align}
where $\mathbf{e}_j = \langle s_{j}, n_{j} \rangle$ is a tuple of the beginning ($s_{j}$) and the end ($n_{j}$) time tag for the segment of $y_j$, and $n_j > s_j $ while $n_j, s_j \in [1, T]$; $y_j \in \mathcal{Y}$ and $\mathcal{Y}$ denotes the vocabulary set; $\bar{\bm x}_j$ is the embedding vector of the segment corresponding to the token $y_j$.  In this case, $Z({\bm X})$ sums over all the possible $({\bm y, \bm E})$ pairs, i.e.,
\begin{eqnarray}
Z({\bm X}) = \sum_{\bm y,\bm E} \prod_{j=1}^J \exp f \left( y_j, {\bm e}_j, \bar{\bm x}_j \right).
\end{eqnarray}
Similar to other CRFs, the function $f(\cdot)$ is defined as
\begin{eqnarray}
\label{eq:phi}
f \left( y_j, {\bm e}_j, \bar{\bm x}_t \right) = \mathbf{w}^\top \Phi (y_j, {\bm e}_j, \bar{\bm x}_j),
\end{eqnarray}
where $\Phi(\cdot)$ denotes the feature function, and $\mathbf{w}$ is the weight vector. Most of conventional approaches for SCRF-based acoustic models use a manually defined feature function $\Phi(\cdot)$, where the features and segment boundary information are provided by an auxiliary system~\cite{zweig2011speech, fosler2013conditional}. In~\cite{kong2015segmental, lu2016segmental}, we proposed an end-to-end training approach for SCRFs, where $\Phi(\cdot)$ was defined with neural networks, and the segmental level features were learned by RNNs. The model was referred to as the segmental RNN (SRNN), and it will be used as the implementation of the SCRF acoustic model for multitask learning in this study.

\subsection{Feature Function and Acoustic Embedding}

\begin{figure}[t]
\small
\centerline{\includegraphics[width=0.4\textwidth]{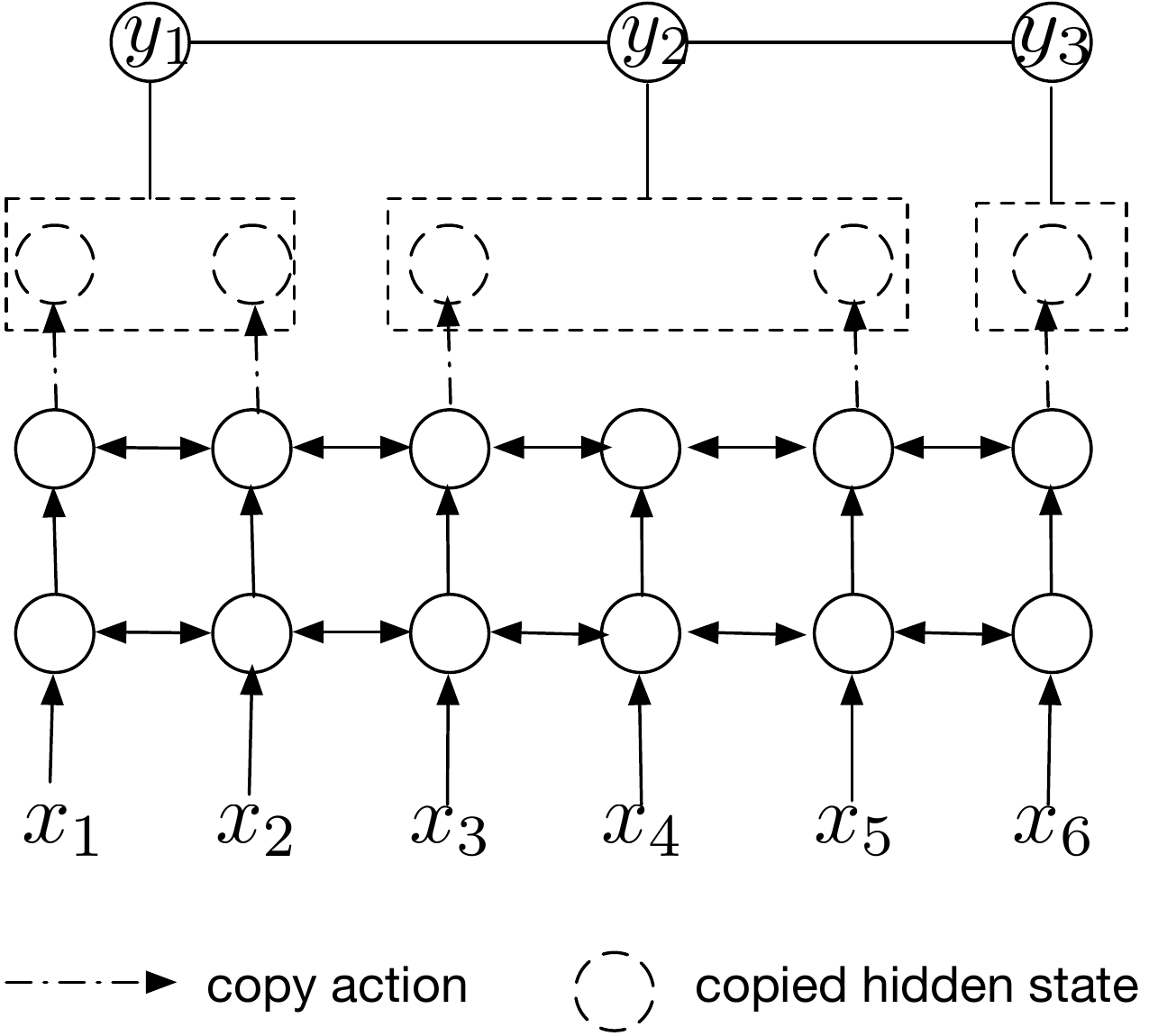}} \vskip -2mm
\caption{A segmental RNN with context-aware embedding. The acoustic segmental embedding vector is composed by the hidden states from the RNN encoder corresponding to the beginning and end time tags. }  
\label{fig:srnn}
\vskip -4mm
\end{figure}

SRNN uses an RNN to learn segmental level acoustic embeddings. Given the input sequence ${\bm X} = ({\bm x}_1, \cdots, {\bm x}_T)$, and we need to compute the embedding vector $\bar{\bm x}_j$ in Eq.~\eqref{eq:phi} corresponding to the segment ${\bm e}_j = \langle s_j, n_j\rangle$. Since the segment boundaries are known, it is straightforward to employ an RNN to map the segment into a vector as 
\begin{align}
\left [ \begin{array}{l}
{\bm h}_{s_j}  \\
{\bm h}_{s_j+1}  \\
\hskip1mm \vdots \\
{\bm h}_{n_j}   
\end{array} \right] = 
\left [ \begin{array}{l}
  \text{\tt RNN}({\bm h}_0, {\bm x}_{s_j}) \\
  \text{\tt RNN}({\bm h}_{s_j}, {\bm x}_{s_j+1}) \\
  \hskip1cm \vdots \\
  \text{\tt RNN}({\bm h}_{n_j-1}, {\bm x}_{n_j}) 
\end{array} \right]
\end{align}
where ${\bm h}_0$ denotes the initial hidden state, which is initialized to be zero. {\tt RNN}($\cdot$) denotes the nonlinear recurrence operation used in an RNN, which takes the previous hidden state and the feature vector at the current timestep as inputs, and produce an updated hidden state vector. Given the recurrent hidden states, the embedding vector can be simply defined as $\bar{\bm x}_j= {\bm h}_{n_j}$ as in our previous work~\cite{lu2016segmental}. However, the drawback of this implementation is the large memory cost, as we need to store the array of hidden states $({\bm h}_{s_j}, \cdots, {\bm h}_{n_j})$ for all the possible segments $\langle s_j, n_j\rangle$. If we denote $H$ as the dimension of an RNN hidden state, the memory cost will be on the order of $O(T^2H)$, where $T$ is the length of $\bm X$. It is especially problematic for the joint model as the CTC model requires additional memory space. In this work, we adopt another approach that requires much less memory. In this approach, we use an RNN to read the whole input sequence as 
\begin{align}
\left [ \begin{array}{c}
{\bm h}_{1}  \\
{\bm h}_{2}  \\
\vdots \\
{\bm h}_{T}   
\end{array} \right] = 
\left [ \begin{array}{l}
  \text{\tt RNN}({\bm h}_0, {\bm x}_{1}) \\
  \text{\tt RNN}({\bm h}_1, {\bm x}_{2}) \\
 \hskip1cm \vdots \\
  \text{\tt RNN}({\bm h}_{T-1}, {\bm x}_{T}) 
\end{array} \right]
\end{align}
and we define the embedding vector for segment ${\bm e} = \langle k, t\rangle$ as
\begin{align}
\bar{\bm x}_j =
\left [ \begin{array}{c}
{\bm h}_{s_j}  \\
{\bm h}_{n_j}   
\end{array} \right]
\end{align}
In this case, we only provide the context information for the feature function $\Phi(\cdot)$ to extract segmental features. We refer this approach as context-aware embedding. Since we only need to read the input sequence once, the memory requirement is on the order of $O(TH)$, which is much smaller. The cost, however, is the slightly degradation of the recognition accuracy. This model is illustrated by Figure \ref{fig:srnn}.
 
The feature function $\Phi(\cdot)$ also requires a vector representation of the label $y_j$. This embedding vector can be obtained using a linear embedding matrix, following common practice for RNN language models. More specifically, $y_j$ is first represented as a one-hot vector ${\bm v}_j$, and it is then mapped into a continuous space by a linear embedding matrix ${\bm M}$ as
\begin{eqnarray}
{\bm u}_j = {\bm M \bm v}_j
\end{eqnarray}
Given the acoustic embedding $\bar{\bm x}_j$ and label embedding $\bm u_j$, the feature function $\Phi(\cdot)$ can be represented as 
\begin{align}
\label{eq:feafunc}
\Phi(y_j, {\bm e}_j, \bar{\bm x}_j) = \sigma({\bm W}_1{\bm u}_j + {\bm W}_2\bar{\bm x}_j + \bm b),
\end{align}
where $\sigma$ denotes a non-linear activation function (e.g., sigmoid or tanh); $\bm W_1, \bm W_2$ and $\bm b$ are weight matrices and a bias vector. Eq. \eqref{eq:feafunc} corresponds to one layer of non-linear transformation. In fact, it is straightforward to stack multiple nonlinear layers in this feature function.

\subsection{Loss Function}
\label{sec:hidden_loss}

For speech recognition, the segmentation labels ${\bm E}$ are usually unknown in the training set. In this case, we cannot train the model directly by maximizing the conditional probability in Eq. \eqref{eq:scrf}. However, the problem can be addressed by marginalizing out the segmentation variable as
\begin{align}
\mathcal{L}_{\mathit{scrf}} &= - \log P({\bm y} \mid {\bm X}) \nonumber \\
 &= - \log \sum_{{\bm E}} P({\bm y, \bm E} \mid {\bm X}) \nonumber \\
&= - \log \underbrace{\sum_{{\bm E}} \prod_j \exp f \left( y_j, {\bm e}_j, \bar{\bm x}_j \right)}_{ Z({\bm X}, \mathbf{y})} + \log Z({\bm X}),
\end{align}
where $Z({\bm X}, {\bm y})$ denotes the summation over all the possible segmentations when only ${\bm y}$ is observed. To simplify notation, the objective function $\mathcal{L}_{\mathit{scrf}}$ is defined here with only one training utterance. 

However, the number of possible segmentations is exponential in the length of ${\bm X}$, which makes the na\"{i}ve computation of both $Z({\bm X}, {\bm y})$ and $Z({\bm X})$ impractical. To address this problem, a dynamic programming algorithm can be applied, which can reduce the computational complexity to $O(T^2\cdot |\mathcal{Y}|)$~\cite{sarawagi2004semi}. The computational cost can be further reduced by limiting the maximum length of all the possible segments. The reader is referred to~\cite{lu2016segmental} for further details including the decoding algorithm.

\section{Connectionist Temporal Classification }

CTC also directly computes the conditional probability $P(\bm y \mid \bm X)$, with the key difference from SCRF in that it normalizes the probabilistic distribution at the frame level. To address the problem of length mismatch between the input and output sequences, CTC allows repetitions of output labels and introduces a special  blank token ($-$), which represents the probability of not emitting any label at a particular time step. The conditional probability is then obtained by summing over all the probabilities of all the paths that corresponding to $\bm y$ after merging the repeated labels and removing the blank tokens, i.e.,
\begin{align} 
P(\bm y \mid \bm X) = \sum_{\bm \pi \in \Psi(\bm y)} P(\bm \pi \mid \bm X),
\end{align}
where $\Psi(\bm y)$ denotes the set of all  possible paths that correspond to $\bm y$ after repetitions of labels and insertions of the blank token. Now the length of $\bm \pi$ is the same as $\bm X$, the probability $P(\bm \pi \mid \bm X)$ is then approximated by the independence assumption as 
\begin{align} 
P(\bm \pi \mid \bm X) \approx \prod_{t=1}^T P(\pi_t \mid \bm x_t),
\end{align}
where $\pi_t $ ranges over $\mathcal{Y}\cup \{-\}$, and $P(\pi_t \mid \bm x_t)$ can be computed using the softmax function. The training criterion for CTC is to maximize the conditional probability of the ground truth labels, which is equivalent to minimizing the negative log likelihood:
\begin{align} 
\mathcal L_{\mathit{ctc}} = -\log P(\bm y \mid \bm X),
\end{align}
which can be reformulated as the CE criterion. More details regarding the computation of the loss and the backpropagation algorithm to train CTC models can be found in~\cite{graves2006connectionist}.

\section{Joint Training Loss}

Training the two models jointly is trivial.  We can simply interpolate the CTC and SCRF loss functions as
\begin{align} 
\label{eq:mtl}
\mathcal L = \lambda \mathcal{L}_{\mathit{ctc}} + (1-\lambda)\mathcal{L}_{\mathit{scrf}},
\end{align}
where $\lambda \in[0, 1]$ is the interpolation weight. The two models share the same neural network for feature extraction. In this work, we focus on the RNN with long short-term memory (LSTM)~\cite{hochreiter1997long} units for feature extraction. Other types of neural architecture, e.g., convolutional neural network (CNN) or combinations of CNN and RNN, may be considered in future work. 

\section{Experiments}

Our experiments were performed on the TIMIT database, and both the SRNN and CTC models were implemented using the DyNet toolkit~\cite{neubig2017dynet}. We followed the standard protocol of the TIMIT dataset, and our experiments were based on the Kaldi recipe~\cite{povey2011kaldi}. We used the core test set as our evaluation set, which has 192 utterances. Our models were trained with 48 phonemes, and their predictions were converted to 39 phonemes before scoring. The dimension of $\mathbf{u}_j$ was fixed to be 64, and the dimension of $\mathbf{w}$ in Eq.~\eqref{eq:phi} is also 64. We set the initial SGD learning rate to be 0.1, and we exponentially decay the learning rate by 0.75 when the validation error stopped decreasing. We also subsampled the acoustic sequence by a factor of 4 using the hierarchical RNN as in~\cite{lu2016segmental}. Our models were trained with dropout regularization~\cite{srivastava2014dropout}, using a specific implementation for recurrent networks~\cite{zaremba2014recurrent}. The dropout rate was 0.2 unless specified otherwise. Our models were randomly initialized with the same random seed.

\begin{table}
 \centering \small
\caption{Phone error rates of baseline CTC and SRNN models. }
\label{tab:baseline}
\begin{tabular}{cl|ccccccc}
\hline

\hline
Model & Features  & \#Layer & Dim & dev  & eval \\ \hline
SRNN & FBANK & 3 & 128 & 19.2 & 20.5 \\
%SRNN & FBANK & 6 & 128 & 18.6 & 20.9 \\
SRNN & fMLLR & 3 & 128 & 17.6 & 19.2 \\
SRNN & FBANK & 3& 250 & 18.1 & 20.0\\
SRNN & fMLLR & 3 & 250 & 16.6 & 17.9 \\ \hline
CTC & FBANK & 3&  128 & 20.0 & 21.8 \\ 
CTC & fMLLR & 3 & 128 & 17.7 & 18.4 \\
CTC & FBANK & 3&  250 & 17.7 & 19.9 \\ 
CTC & fMLLR & 3 & 250 & 16.7 & 17.8 \\ \hline

\hline
\end{tabular}
\end{table}

\begin{table}
 \centering \small
\caption{Results of three types of acoustic features. }
\label{tab:mtl}
\begin{tabular}{ll|cccccc}
\hline

\hline
Model & Features   & Dim & dev  & eval \\ \hline
SRNN & FBANK & 250 & 18.1 & 20.0 \\
$\quad$+MTL & FBANK& 250 &17.5& 18.7 \\
SRNN & fMLLR & 250 & 16.6 & 17.9 \\
$\quad$+MTL  & fMLLR & 250 & 15.9 & 17.5 \\ \hline
CTC & FBANK & 250 & 17.7 & 19.9\\ 
$\quad$+MTL & FBANK& 250 &  17.2 & 18.9 \\
CTC & fMLLR & 250 & 16.7 & 17.8 \\ 
$\quad$+MTL   & fMLLR & 250 & 16.2 & 17.4 \\ \hline

\hline
\end{tabular}
\vskip-2mm
\end{table}

\begin{figure*}[t]
\small
\centerline{\includegraphics[width=1\textwidth]{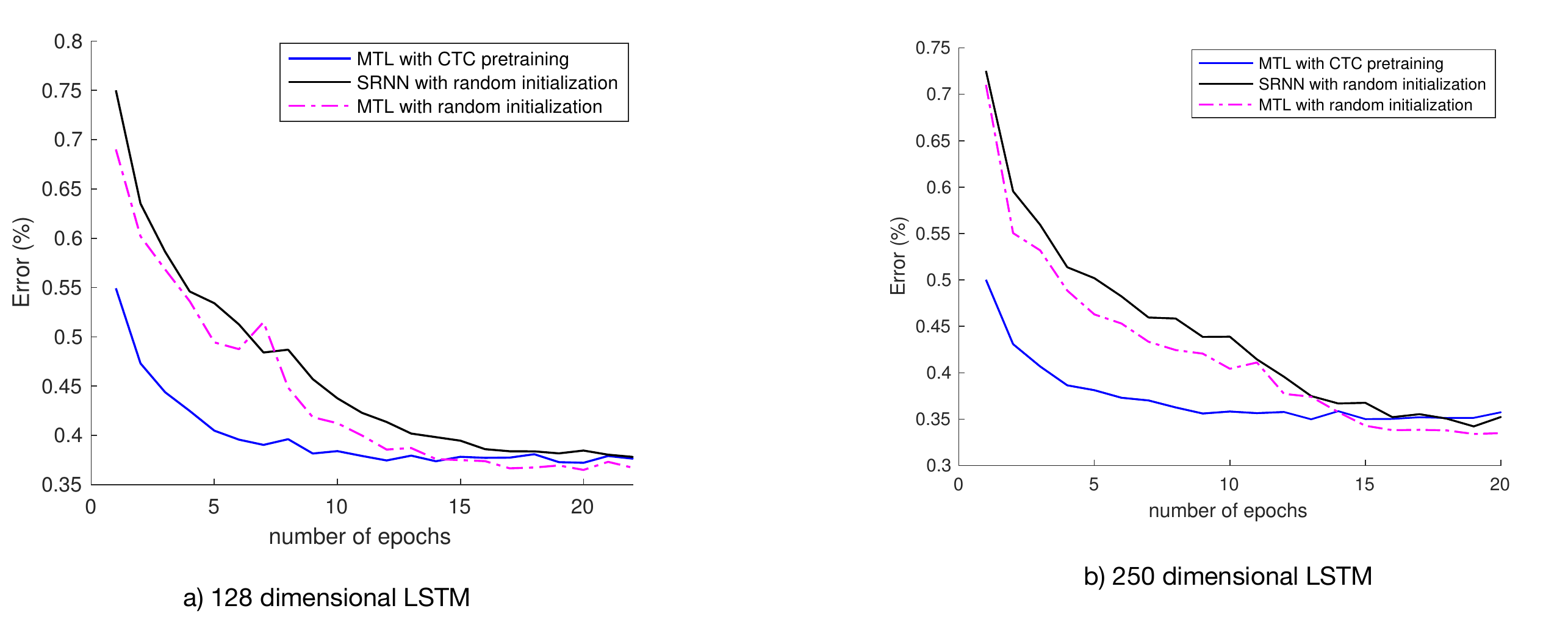}} \vskip -2mm
\caption{Convergence curves with and without CTC pretraining in multitask learning framework.}  
\label{fig:ctc-pretrain}
\vskip -2mm
\end{figure*}

\subsection{Baseline Results}

Table \ref{tab:baseline} shows the baseline results of SRNN and CTC models using two different kinds of features. The FBANK features are 120-dimensional with delta and delta-delta coefficients, and the fMLLR features are 40-dimensional, which were obtained from a Kaldi baseline system. We used a 3-layer bidirectional LSTMs for feature extraction, and we used the greedy best path decoding algorithm for both models. Our SRNN and CTC achieved comparable phone error rate (PER) for both kinds of features. However, for the CTC system, Graves et al.~\cite{graves2013speech} obtained a better result, using about the same size of neural network (3 hidden layers with 250 hidden units of bidirectional LSTMs), compared to ours (18.6\% vs. 19.9\%). Apart from the implementation difference of using different code bases, Graves et al.~\cite{graves2013speech} applied the prefix decoding with beam search, which may have lower search error than our best path decoding algorithm. 

\subsection{Multitask Learning Results}

Table~\ref{tab:mtl} shows results of multitask learning for CTC and SRNN using the interpolated loss in Eq.~\eqref{eq:mtl}. We only show results of using LSTMs with 250 dimensional hidden states. The interpolation weight was set to be 0.5. In our experiments, tuning the interpolation weight did not further improve the recognition accuracy. From Table~\ref{tab:mtl}, we can see that multitask learning improves recognition accuracies of both SRNN and CTC acoustic models, which may due to the regularization effect of the joint training loss. The improvement for FBANK features is much larger than fMLLR features. In particular, with multitask learning, the recognition accuracy of our CTC system with best path decoding is comparable to the results obtained by Graves et al.~\cite{graves2013speech} with beam search decoding. 

One of the major drawbacks of SCRF models is their high computational cost. In our experiments, the CTC model is around 3--4 times faster than the SRNN model that uses the same RNN encoder. The joint model by multitask learning is slightly more expensive than the stand-alone SRNN model. To cut down the computational cost, we investigated if CTC can be used to pretrain the RNN encoder to speed up the training of the joint model. This is analogous to sequence training of HMM acoustic models, where the network is usually pretrained by the frame-level CE criterion. Figure~\ref{fig:ctc-pretrain} shows the convergence curves of the joint model with and without CTC pretraining, and we see  pretraining indeed improves the convergence speed of the joint model. %For sequence training of HMMs, it usually takes around 4 - 6 iterations to converge.  

\section{Conclusion}

We investigated multitask learning with CTC and SCRF for speech recognition in this paper. Using an RNN encoder for feature extraction, both CTC and SCRF can be trained end-to-end, and the two models can be trained together by interpolating the two loss functions. From experiments on the TIMIT dataset, the multitask learning approach improved the recognition accuracies of both CTC and SCRF acoustic models. We also showed that CTC can be used to pretrain the RNN encoder, speeding up the training of the joint model. In the future, we will study the multitask learning approach for larger-scale speech recognition tasks, where the CTC pretraining approach may be more helpful to overcome the problem of high computational cost.

\section{Acknowledgements}

We thank the NVIDIA Corporation for the donation of a Titan X GPU.

\bibliographystyle{IEEEtran}
\bibliography{bibtex}

\end{document}